\documentclass[10pt]{article} 
\usepackage[accepted]{tmlr}

\usepackage[hidelinks,hypertexnames=false]{hyperref}
\usepackage{url}
\usepackage{amsmath,amssymb,amsthm,mathtools,amsfonts,bm}
\usepackage{booktabs}
\usepackage[table]{xcolor}
\usepackage{colortbl}
\usepackage{multirow}
\usepackage{graphicx}
\usepackage[section]{placeins}
\usepackage{algorithm}
\usepackage{algpseudocode}
\usepackage{pifont}

\graphicspath{{figures/}}
\colorlet{ours}{blue!6}
\definecolor{revblue}{rgb}{0,0,1}

\newtheorem{proposition}{Proposition}

\newcommand{\srank}{r_{\mathrm{stable}}}

\newcommand{\rlvr}{\textsc{rlvr}}
\newcommand{\sft}{\textsc{sft}}
\newcommand{\ppo}{\textsc{ppo}}
\newcommand{\grpo}{\textsc{grpo}}
\newcommand{\dpo}{\textsc{dpo}}

\newcommand{\revised}[1]{#1}
\newenvironment{revisedblock}{\begingroup}{\endgroup}
\ifdefined\NoBlue
  
\else
  
\fi

\def\eqref#1{equation~\ref{#1}}

\def\1{\bm{1}}

\DeclareMathAlphabet{\mathsfit}{\encodingdefault}{\sfdefault}{m}{sl}
\SetMathAlphabet{\mathsfit}{bold}{\encodingdefault}{\sfdefault}{bx}{n}

\title{Spectral Rank and Step-Direction Dynamics in Language-Model Post-Training}

\author{
\name Khurram Khalil
\email khurram.khalil@missouri.edu \\
\addr Department of Electrical Engineering \& Computer Science\\
University of Missouri-Columbia, MO, USA
\AND
\name Khaza Anuarul Hoque
\email hoquek@missouri.edu \\
\addr Department of Electrical Engineering \& Computer Science\\
University of Missouri-Columbia, MO, USA
}

\def\month{November}
\def\year{2026}
\def\openreview{\url{https://openreview.net/forum?id=UGIcayxHMI}}

\begin{document}

\maketitle

\begin{abstract}
Understanding how language models change during post-training requires characterizing both the structure of cumulative weight updates and the directions of successive training displacements. Existing studies report seemingly contrasting spectral properties, including nearly full-rank updates and low stable-rank cumulative changes. We show that these observations are mathematically compatible and investigate how weight-update directions evolve during training. For singular values $\sigma_i\propto i^{-\alpha}$, $0<\alpha\leq 1/2$, we prove that stable rank becomes negligible relative to fixed-energy rank as dimension grows. \revised{Full singular value decompositions of selected OLMo-2-7B updates verify this separation. Three attention updates from proximal policy optimization (PPO) have stable ranks $9.6$--$18.3$ but require $532$--$1{,}222$ components for $80\%$ of their energy. We then measure consecutive checkpoint-displacement cosines using full-model inner products. Across nine public stages of reinforcement learning from verifiable rewards (RLVR), all $131$ adjacent pairs have negative cosine on their native checkpoint grids, including PPO and Group Relative Policy Optimization (GRPO) releases. Supervised comparisons across four model families reveal configuration-dependent alignment, with negative late-training cosines in Mistral and FP32 Llama runs. A controlled OLMo-2-1B study with three paired seeds and a $2{,}048$-token response cap compares group-relative advantages with a learned critic and generalized advantage estimation. At matched 60-rollout windows, their late-training mean cosines are $+0.259$ and $+0.012$, with a paired difference of $-0.247$ and a seed-level $95\%$ interval of $[-0.358,-0.137]$.} Together, these results separate spectral concentration, energy coverage, and trajectory direction, and show that the advantage system affects directional alignment under matched training conditions.
\end{abstract}

\section{Introduction}
\label{sec:intro}

Language-model post-training changes a model's behavior by changing its weights. Supervised fine-tuning (\sft{}) learns from demonstration responses, while reinforcement learning from verifiable rewards (\rlvr{}) learns from sampled responses scored by an automatic checker. Weight-space measurements describe how these procedures alter a pretrained model and inform methods that operate on weight differences, including task arithmetic and low-rank adaptation \citep{ilharco2023editing,hu2022lora}. Interpreting those measurements requires separating the mathematical object being measured from the functional applied to it.

That distinction matters when comparing recent accounts of post-training geometry. \citet{mukherjee2025rl} report sparse, nearly full-algebraic-rank \rlvr{} updates. \citet{zhu2025path} analyze update localization, numerical precision, and alignment with principal weight subspaces. \citet{shen2026opd} compare cumulative updates in on-policy distillation, \rlvr{}, and \sft{}, reporting low stable rank for distillation and the \rlvr{} endpoint. These studies characterize different aspects of adaptation. Yet their rank descriptions can appear incompatible if algebraic rank, stable rank, and energy coverage are treated as interchangeable. A comparison on the same update matrix is needed to determine what each description actually implies. We address this question with an explicit spectral construction and full-spectrum measurements of selected updates.

Even a complete endpoint spectrum leaves a second question unanswered. It does not determine whether consecutive training displacements reinforce or oppose one another. Analyses of full weight matrices \citep{he2025poststruct} and trajectory-level linear trends \citep{wang2026linear} provide complementary information, but neither an endpoint change nor a fitted trend determines the inner products between consecutive finite displacements. We therefore measure direction at specified checkpoint widths and separations. These measurements describe local alignment on the saved checkpoint grid.

\begin{revisedblock}
Differences in trajectory alignment raise a third question. How does the advantage estimator affect the direction of weight updates under a shared training schedule? We investigate this question through a controlled comparison of group-relative advantages and a learned critic with generalized advantage estimation. Holding the actor schedule and sampled prompt stream fixed connects the observed geometry to a specific change in the training procedure. 

These investigations provide complementary views of post-training geometry. Spectral analysis characterizes how the cumulative weight change is distributed across singular directions, while trajectory analysis examines how successive weight changes interact during training. The controlled comparison then investigates whether changing the advantage system alters these directional relationships. Together, these measurements distinguish the structure of the final update from the geometry of the path taken to reach it.
\end{revisedblock}

Our contributions follow these three questions. First, we prove that low stable rank and large fixed-energy rank coexist for a family of power-law singular spectra, and verify the separation with full decompositions of OLMo-2-7B cumulative updates. \revised{Second, we document negative consecutive-displacement cosines across nine public \rlvr{} stage trajectories, including \ppo{} and \grpo{} releases, quantify their temporal extent and opposing components, and test supervised alignment across four model families and two Llama precision settings. Third, we identify an advantage-system effect in six matched OLMo-2-1B runs. The critic--GAE arm reduces late-training alignment by $0.247$ on average relative to the group-relative arm, with a paired seed-level $95\%$ interval of $[-0.358,-0.137]$ for the signed difference. Together, these contributions reconcile spectral descriptions of post-training and establish trajectory direction as a separately measurable aspect of adaptation.}

\section{Background and Geometric Quantities}
\label{sec:background}

\begin{revisedblock}
\subsection{How post-training changes the weights}
\label{sec:objectives}

Let $\pi_\theta(y\mid x)$ denote the probability of response $y$ to prompt $x$ under parameters $\theta$. For a batch $\mathcal B$ of tokenized demonstrations $z$, \sft{} minimizes a masked next-token negative log likelihood
\begin{equation}
 L_{\mathrm{SFT}}(\theta)
 =-\frac{\sum_{z\in\mathcal B}\sum_{j=2}^{|z|}m_j(z)
 \log\pi_\theta(z_j\mid z_{<j})}
 {\sum_{z\in\mathcal B}\sum_{j=2}^{|z|}m_j(z)}.
 \label{eq:sft}
\end{equation}
Here $m_j(z)$ selects supervised positions and excludes padding. Response-only supervision selects response positions; our \sft{} experiments supervise all non-padding conversation tokens with a preceding token, including prompt and response text. The target tokens are fixed demonstrations. The loss supplies gradients, and an optimizer such as Adam converts those gradients into parameter changes. A saved weight displacement therefore aggregates the optimizer-preconditioned updates between checkpoints.

In \rlvr{}, the current policy samples responses and a checker supplies a reward $R(x,y)$, for example whether a final mathematical answer is correct. A common objective balances expected reward against deviation from a frozen reference policy $\pi_{\mathrm{ref}}$. Practical implementations optimize sample-based surrogate losses. Proximal policy optimization (\ppo{}) \citep{schulman2017proximal} uses a token probability ratio $r_j(\theta)=\pi_\theta(y_j\mid x,y_{<j})/\pi_{\mathrm{old}}(y_j\mid x,y_{<j})$ and the clipped objective
\begin{equation}
 L_{\mathrm{actor}}(\theta)
 =-\mathbb E_j\min\left(r_j(\theta)\widehat A_j,
 \operatorname{clip}(r_j(\theta),1-\epsilon,1+\epsilon)\widehat A_j\right)
 +\beta\widehat D_{\mathrm{KL}}.
 \label{eq:actor}
\end{equation}
Here $\widehat A_j$ estimates the advantage of a sampled token or response, $\epsilon$ controls ratio clipping, and $\beta$ weights the reference-policy penalty. Multiple optimization passes may reuse the same sampled responses.

The advantage estimator distinguishes important implementations of this objective. A learned value function estimates expected future reward, and generalized advantage estimation (GAE) combines temporal-difference residuals \citep{schulman2016gae}. Group Relative Policy Optimization (\grpo{}) instead compares rewards among responses to the same prompt without a learned value function \citep{shao2024deepseekmath}. Our controlled arms share Equation~\ref{eq:actor} and differ in this advantage-system package. We name the two arms \emph{group-relative} and \emph{critic--GAE} after their advantage estimators. Appendix~\ref{app:protocol} specifies their estimators and training configuration.

The reference for a weight displacement must follow this training sequence. The public OLMo-2-7B instruction model follows \sft{}, direct preference optimization (\dpo{}), and then \rlvr{} \citep{olmo2025olmo2}. \dpo{} learns from preferred and dispreferred responses \citep{rafailov2023direct}. Consequently, the pure \rlvr{} displacement is $W_{\mathrm{RLVR}}-W_{\mathrm{DPO}}$. Subtracting the \sft{} checkpoint instead adds the intervening \dpo{} change. All PPO endpoint measurements use the immediate \dpo{} initialization.
\end{revisedblock}

\subsection{Endpoint functionals}
\label{sec:functionals}

For a nonzero matrix displacement $\Delta=W_T-W_0\in\mathbb R^{m\times n}$, let $\sigma_1\geq\cdots\geq\sigma_d\geq0$, $d=\min(m,n)$, be its singular values. Algebraic rank counts nonzero singular values. Stable rank measures total squared singular energy relative to the largest squared singular value
\begin{equation}
 \srank(\Delta)=\frac{\|\Delta\|_F^2}{\|\Delta\|_2^2}
 =\frac{\sum_i\sigma_i^2}{\sigma_1^2},\qquad
 E_k(\Delta)=\frac{\sum_{i=1}^k\sigma_i^2}{\sum_{i=1}^d\sigma_i^2},\qquad
 k_\tau(\Delta)=\min\{k\mid E_k(\Delta)\geq\tau\}.
 \label{eq:ranks}
\end{equation}
We call $k_\tau$ the \emph{energy-threshold rank}. This avoids ambiguity with entropy-based effective rank \citep{roy2007effective}. \revised{Stable rank is a continuous measure of spectral concentration, while $k_{0.8}$ explicitly counts the components needed for $80\%$ of the squared Frobenius norm. These definitions imply $1\leq\srank\leq\operatorname{rank}(\Delta)$ and $k_\tau\geq\tau\srank$, since $\sum_{i=1}^k\sigma_i^2\leq k\sigma_1^2$. They supply no converse bound of comparable size, as the next proposition shows.}

Intuitively, smaller stable rank indicates stronger concentration around the largest singular value, whereas energy-threshold rank measures how many singular components are needed to capture a specified fraction of the total energy. For example, an update may contain one dominant singular value and hundreds of smaller but collectively significant singular values. Such an update can have low stable rank while requiring hundreds of components to capture a specified energy threshold (e.g., 80\%, as used in our experiments).

\begin{proposition}[Separation of stable rank and fixed-energy rank]
\label{prop:decouple}
For $\sigma_i=i^{-\alpha}$, $i=1,\ldots,n$, and fixed $\tau\in(0,1)$, $\srank/k_\tau\to0$ as $n\to\infty$ whenever $0<\alpha\leq1/2$. For $0<\alpha<1/2$, with $b=1-2\alpha$,
\begin{equation}
 \srank\leq 1+\frac{n^b-1}{b},\qquad
 k_\tau\geq \tau^{1/b}(n^b-1)^{1/b}=\Theta(n).
 \label{eq:powerlaw}
\end{equation}
For $\alpha=1/2$, $\srank\leq1+\log n$ and $k_\tau\geq n^\tau/e$.
\end{proposition}

Appendix~\ref{app:proof} proves the proposition by integral comparison. The construction shows how a leading singular value and a broad energetic tail produce the two rank descriptions simultaneously.

These spectral quantities do not specify how many entries of an update are zero. We therefore consider support sparsity separately. Exact zero-entry fraction $1-\|\Delta\|_0/(mn)$ and thresholded sparsity depend on the coordinate basis and numerical precision. A diagonal matrix with all diagonal entries nonzero has fixed support and an arbitrary positive singular spectrum. It can have stable rank close to one or equal to $n$ without changing that support. Orthogonal changes of basis preserve its spectrum while generally making it dense. Thus support sparsity does not determine either spectral rank.

\begin{revisedblock}
\subsection{Finite-displacement direction}
\label{sec:direction}

Endpoint functionals do not specify the path to the endpoint. Let $\theta_t\in\mathbb R^P$ be the vector obtained by flattening and concatenating the model parameters at checkpoint $t$, with tied parameters counted once; $P$ is the number of parameter coordinates measured. On a grid with width $h$, define the full-parameter displacement $D_j^{(h)}=\theta_{jh}-\theta_{(j-1)h}$. Thus $D_j^{(h)}\in\mathbb R^P$. We use the Euclidean inner product $\langle x,y\rangle=x^\top y$ and norm $\|x\|_2=\sqrt{x^\top x}$. For nonzero displacements, define their directional correlation at separation $\ell$ as
\begin{equation}
 \rho_{j,\ell}^{(h)}=
 \frac{\langle D_j^{(h)},D_{j-\ell}^{(h)}\rangle}
 {\|D_j^{(h)}\|_2\|D_{j-\ell}^{(h)}\|_2}.
 \label{eq:cosine}
\end{equation}

A positive cosine indicates that the compared displacements point in broadly similar directions, whereas a negative cosine indicates that they contain opposing directional components. Importantly, negative cosine does not imply that training is reversing or that model performance is deteriorating. Instead, it describes only the geometric relationship between the compared weight changes. For consecutive displacements, $\ell=1$. Width $h$ and separation $\ell$ are distinct. Increasing $h$ aggregates more training updates, whereas increasing $\ell$ compares displacements further apart on a fixed grid. We refer to a negative consecutive cosine as \emph{consecutive anti-alignment}.

The amount projected against the previous displacement also depends on norms. Writing $D_j=a_jD_{j-1}+D_j^\perp$, with $\langle D_j^\perp,D_{j-1}\rangle=0$, gives
\begin{equation}
 a_j=\frac{\langle D_j,D_{j-1}\rangle}{\|D_{j-1}\|_2^2}
 =\rho_{j,1}^{(h)}\frac{\|D_j\|_2}{\|D_{j-1}\|_2}.
 \label{eq:projection}
\end{equation}
For example, cosine $-0.25$ gives $a_j=-0.025$ if the current norm is one tenth of the previous norm, and $a_j=-0.5$ if it is twice as large.
\end{revisedblock}

\section{Data and Measurement Protocol}
\label{sec:setup}

\begin{revisedblock}
\subsection{Public trajectories and supervised comparisons}
\label{sec:publicsetup}

Our public-release analysis covers nine \rlvr{} stages in the OLMo-2, OLMoE, and Olmo-3 releases. Table~\ref{tab:production} lists each stage and its native checkpoint width. The OLMo-2 stages include 7B and 13B \ppo{} releases and 1B and 32B \grpo{} releases. We report successive stages within each model lineage separately and retain their shared lineage in the checkpoint inventory.

For \sft{}, we analyze author-run trajectories from four model families on a uniform 500-optimizer-step grid through step 5,000. The comparison includes OLMo-2-7B and Llama-3.2-3B at seeds 42, 43, and 44, Mistral-7B-v0.3 at seeds 42 and 43, and Qwen2.5-3B at seed 42. Each trajectory contains ten displacements and nine adjacent pairs. All runs use the T\"ulu-3 OLMo-2 data mixture \citep{lambert2024tulu3,olmo2025olmo2}, the same plain-text role formatting and all-token objective in Equation~\ref{eq:sft}, effective batch size 64, sequence cap 4,096, and learning rate $10^{-5}$ with 3\% warmup and linear decay over the two-epoch horizon. Training stops at step 5,000 while preserving that schedule. The OLMo seed-42 run continues to step 29,356 and supplies the \sft{} endpoint for spectral analysis.

The cross-family runs store trainable parameters and Adam states in BF16. Two additional Llama runs at seeds 42 and 43 use FP32 parameters and Adam states with BF16 autocast, retaining the data mixture, effective batch size, learning-rate schedule, and measurement grid. Appendix~\ref{app:e4} reports each completed run, the precision comparison, and the full-model measurement checks. These comparisons match example and optimizer-step budgets; tokenization and realized token counts vary across families.

The spectral validation uses fourteen matrices selected by layer position and shape before inspecting approximation errors. For each of the 7B PPO and author-run 7B \sft{} endpoints, we select query projections at layers 0, 16, and 31, up projections at layers 0 and 31, the down projection at layer 16, and the embedding matrix. Table~\ref{tab:dimensions} states their dimensions and the complete 7B matrix inventory. The endpoints differ in objective, initialization, data exposure, and horizon. We examine each matrix's spectral functionals and compare their concentration and energy coverage across these endpoints.
\end{revisedblock}

\begin{table}[t]

\centering
\caption{Matrix dimensions in the OLMo-2-7B endpoint analysis. Stored orientation is rows by columns. The 226 matrices exclude one-dimensional normalization weights, which remain included in full-model directional measurements.}
\label{tab:dimensions}
\begin{tabular}{lrrl}
\toprule
Matrix class & Rows & Columns & Count \\
\midrule
Attention query, key, value, output & 4,096 & 4,096 & 128 \\
MLP up and gate & 11,008 & 4,096 & 64 \\
MLP down & 4,096 & 11,008 & 32 \\
Token embedding and language-model head & 100,352 & 4,096 & 2 \\
\bottomrule
\end{tabular}
\end{table}

\begin{revisedblock}
\subsection{Controlled advantage-system comparison}
\label{sec:matchedsetup}

The controlled study starts both arms from OLMo-2-0425-1B-DPO and trains on RLVR-GSM for 360 rollout batches. Seeds 42, 43, and 44 each define a paired prompt stream. Both arms use eight prompts and eight sampled responses per prompt, four update passes, eight-response minibatches, learning rate $4\times10^{-7}$ with linear decay, ratio clipping $0.2$, KL coefficient $0.05$, and prompt and response caps of 2,048 tokens. This response cap follows the published OLMo-2 PPO recipe. The rollout batch is smaller than that published recipe, and the experiment uses a shared custom trainer. The shared trainer holds the actor schedule fixed while changing the advantage system.

Each rollout batch generates 64 responses and performs 32 actor optimizer updates. A displacement width of 60 rollouts therefore represents 3,840 responses and 1,920 actor optimizer updates. We match these quantities between arms and report realized generated-token counts in Appendix~\ref{app:budgets}. Generated-token totals vary with the evolving policies. Public trajectories retain the release-specific step units listed in Table~\ref{tab:production}.

The prespecified primary outcome is each seed's mean consecutive cosine at width 60 for endpoints after rollout 180, comprising endpoints 240, 300, and 360. We report all three seed values and the paired critic-minus-group difference. A Student interval uses the three seed-level differences with two degrees of freedom. Widths 10, 20, and 40 and all-window summaries are secondary descriptive checks. Appendix~\ref{app:protocol} records the full protocol.

\subsection{Exact directional measurements and spectral validation}
\label{sec:numerics}

Directional measurements accumulate full-parameter dot products and squared norms in FP64 over coordinate chunks, including normalization weights. For eight retained trajectories, we independently compute the displacement Gram matrix $G_{ij}=\langle D_i,D_j\rangle$, which supplies both higher-separation correlations and norm-aware projections. Online and saved-checkpoint cosines for the six matched runs agree within $1.4\times10^{-12}$ in absolute value.

For the fourteen selected spectral cases, we compute a full FP64 singular decomposition. Reduced QR followed by full SVD preserves the complete singular spectrum of high-aspect-ratio matrices. Leading singular-triplet residuals are below $10^{-10}$. We compare these results with the original randomized SVD configuration, ten oversamples and four power iterations \citep{halko2011finding}, using seeds 42, 43, and 44, and a stronger setting with forty oversamples and eight iterations at seed 42. Rank-256 and rank-512 approximations are computed independently. We separate singular-value error from error introduced by the original FP32 norm and denominator regularizer.

Energy-threshold reporting requires sufficient spectral accuracy. If an approximate top-$K$ energy remains below $\tau$, approximation error alone may explain the shortfall. Without a certified error bound, we label the threshold \emph{unresolved at the computed budget}, not $k_\tau>K$. All energy-threshold ranks used as quantitative evidence below come from full decompositions. Appendix~\ref{app:corrections} documents the corrected initialization and coordinate-sketch errors in the earlier analysis.
\end{revisedblock}

\section{Results}
\label{sec:results}

\begin{figure}[t]

\centering
\includegraphics[width=.8\linewidth]{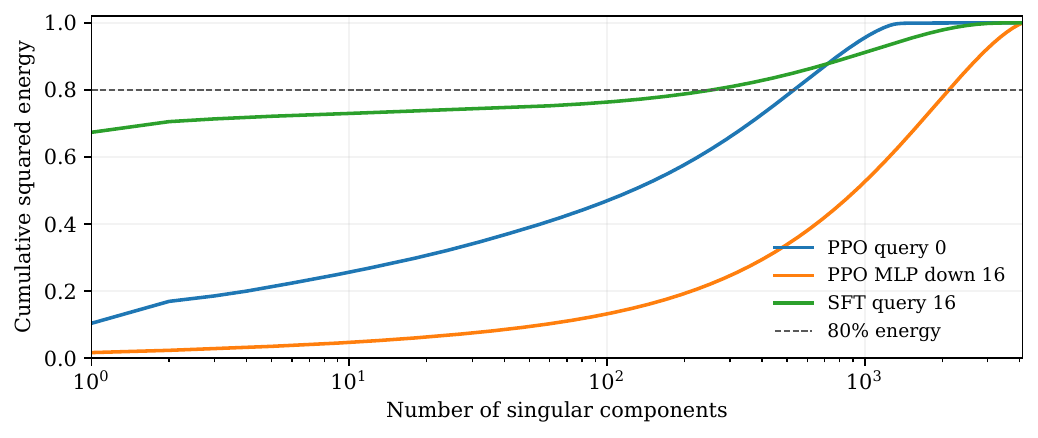}
\caption{Cumulative singular energy from full FP64 decompositions. The PPO query update at layer 0 has stable rank $9.58$ but requires 532 components for $80\%$ energy. The layer-16 MLP down update requires 2,120. The SFT layer-16 query curve also illustrates a threshold near the rank-256 budget.}
\label{fig:energy}
\end{figure}

\begin{revisedblock}
\subsection{Low stable rank coexists with large energy rank}
\label{sec:rankresults}

Full decompositions verify the spectral separation on the same update matrices. Table~\ref{tab:fullranks} reports all fourteen cases, and Figure~\ref{fig:energy} shows representative cumulative energy curves. The three PPO query updates have stable ranks $9.58$, $18.14$, and $18.27$, while their $80\%$-energy ranks are $532$, $1{,}222$, and $1{,}149$. These matrices combine strong leading-value concentration with an energetic tail spanning hundreds of components.

The same distinction applies to the MLP updates. The selected PPO down projection at layer 16 has stable rank $58.78$ but $k_{0.8}=2{,}120$. Its \sft{} counterpart has stable rank $792.32$ and $k_{0.8}=2{,}250$. The stable ranks differ by more than an order of magnitude while their energy ranks are close. Conversely, the \sft{} query update at layer 0 has $k_{0.8}=1$. The two functionals thus reveal different aspects of variation across matrices, with stable rank distinguishing concentration even when energy ranks are similar.

These observations reconcile the logical content of the prior rank descriptions. Near-full algebraic rank permits a dominant leading singular value, and a dominant leading value permits a long energetic tail. Proposition~\ref{prop:decouple} supplies an explicit family with this behavior. Thus, low stable rank alone does not generally imply that a small number of singular components captures most of the update energy.
\end{revisedblock}

\begin{table}[t]

\centering
\caption{Full-SVD endpoint measurements for the shape-stratified 7B sample. PPO subtracts the immediate DPO checkpoint. SFT subtracts the base checkpoint. Each row refers to the same named matrix in the two endpoints. $k_{0.8}$ is computed from the complete spectrum. $E_{512}$ abbreviates $E_k(\Delta)$ from Equation~\ref{eq:ranks} at $k=512$, the fraction of squared singular energy in the leading 512 components.}
\label{tab:fullranks}
\small
\begin{tabular}{lrrrrrr}
\toprule
& \multicolumn{3}{c}{PPO} & \multicolumn{3}{c}{SFT} \\
\cmidrule(lr){2-4}\cmidrule(lr){5-7}
Matrix & $\srank$ & $E_{512}$ & $k_{0.8}$ & $\srank$ & $E_{512}$ & $k_{0.8}$ \\
\midrule
Query 0 & 9.58 & 0.790 & 532 & 1.10 & 0.987 & 1 \\
MLP up 0 & 115.60 & 0.368 & 2,077 & 356.95 & 0.289 & 2,221 \\
Query 16 & 18.14 & 0.536 & 1,222 & 1.48 & 0.848 & 255 \\
MLP down 16 & 58.78 & 0.345 & 2,120 & 792.32 & 0.277 & 2,250 \\
Query 31 & 18.27 & 0.564 & 1,149 & 1.99 & 0.779 & 599 \\
MLP up 31 & 45.72 & 0.342 & 2,114 & 200.56 & 0.291 & 2,210 \\
Embedding & 1026.35 & 0.200 & 2,831 & 916.65 & 0.175 & 2,982 \\
\bottomrule
\end{tabular}
\end{table}

\begin{revisedblock}
\subsection{Production anti-alignment extends beyond PPO}
\label{sec:publicresults}

Every native-grid adjacent pair in the nine public stage trajectories has negative full-model cosine. Table~\ref{tab:production} reports 131 such pairs, with stage means between $-0.263$ and $-0.193$. Consecutive anti-alignment is consistent across this collection of PPO and GRPO releases, model sizes, and checkpoint widths.

The supervised comparison reveals configuration-dependent alignment across four model families. Table~\ref{tab:e4} reports all seeds on the common 500-step grid. The BF16 OLMo, Llama, and Qwen trajectories have positive cosines throughout, with seed means between $+0.2226$ and $+0.2381$. Both Mistral seeds instead have seven negative pairs out of nine and means of $-0.0100$ and $-0.0099$. The Mistral runs thus provide a same-size-class comparison in which \sft{} also exhibits consecutive anti-alignment.

The Llama precision comparison further shows that alignment depends on the numerical training configuration. With FP32 parameters and Adam states, the seed-42 and seed-43 means fall to $+0.0206$ and $+0.0209$, and the final five pairs are negative in both runs. Table~\ref{tab:e4precision} reports the paired series. Positive alignment therefore replicates across several BF16 settings, while the Mistral and FP32 Llama results establish that cosine sign varies within \sft{} itself.

\end{revisedblock}

\begin{table}[t]

\centering
\caption{Consecutive full-model cosines on native public checkpoint grids. Width is in release step labels. Every listed pair is negative. Counts refer to adjacent pairs within each stage. Successive stages retain their shared model lineage.}
\label{tab:production}
\small
\begin{tabular}{llrrr}
\toprule
Public stage & Released method & Width & Pairs & Mean cosine \\
\midrule
OLMo-2 7B & PPO & 60 & 5 & $-0.2202$ \\
OLMo-2 13B RLVR1 & PPO & 60 & 12 & $-0.2446$ \\
OLMo-2 13B RLVR2 & PPO & 60 & 12 & $-0.2541$ \\
OLMo-2 13B final & PPO & 60 & 12 & $-0.2582$ \\
OLMo-2 1B RLVR1 & GRPO & 200 & 12 & $-0.2293$ \\
OLMo-2 1B final & GRPO & 200 & 12 & $-0.2327$ \\
OLMo-2 32B & GRPO & 20 & 16 & $-0.1929$ \\
OLMoE 1B active / 7B & RLVR & 40 & 43 & $-0.2633$ \\
Olmo-3 7B & OlmoRL & 50 & 7 & $-0.2153$ \\
\bottomrule
\end{tabular}
\end{table}

\begin{revisedblock}
Higher-separation correlations quantify how far the directional relationship extends along the trajectory. Table~\ref{tab:lags} reports these correlations for the public 7B PPO trajectory at a fixed width of 60 release steps. Mean cosine changes from $-0.2202$ at separation one to $-0.0244$ and $-0.0087$ at separations two and three. The negative correlation is strongest between adjacent displacements and approaches zero at the larger separations measured here.

Cosine describes directional alignment but does not measure the magnitude of the component opposing the previous displacement. The projection coefficient accounts for the ratio of displacement norms. In the same public 7B run, Equation~\ref{eq:projection} yields coefficients from $-0.2204$ to $-0.1741$. Each displacement has a component opposite the preceding displacement with magnitude 17--22\% of the preceding norm. These observations demonstrate that consecutive weight displacements can exhibit systematic directional opposition, even though their cumulative effect is not characterized by the individual displacement cosines.
\end{revisedblock}

\begin{table}[t]

\centering
\caption{Directional correlations at fixed displacement width. Entries are means with pair counts in parentheses. Separation changes within each row; width does not. Each trajectory retains its native width and training exposure. All six controlled runs appear in Appendix~\ref{app:temporal}.}
\label{tab:lags}
\begin{tabular}{llrrr}
\toprule
Trajectory & Width and unit & $\ell=1$ & $\ell=2$ & $\ell=3$ \\
\midrule
Public 7B PPO & 60 release steps & $-0.2202$ (5) & $-0.0244$ (4) & $-0.0087$ (3) \\
7B SFT seed 42 & 500 trainer steps & $+0.2275$ (9) & $+0.1740$ (8) & $+0.1411$ (7) \\
\bottomrule
\end{tabular}
\end{table}

\begin{revisedblock}
\subsection{A controlled effect on alignment magnitude}
\label{sec:matchedresults}

To test whether the advantage system changes alignment under a shared training schedule, we turn from the public trajectories to the matched study. Table~\ref{tab:matched} reports all three paired seeds for the primary late-training outcome. The group-relative mean is $+0.259$ and the critic--GAE mean is $+0.012$. The paired critic-minus-group difference is $-0.247$, with a seed-level $95\%$ Student interval of $[-0.358,-0.137]$. Every paired seed has lower alignment in the critic arm. The paired contrast shows that the advantage system changes directional alignment consistently across the three seeds.
\end{revisedblock}

\begin{table}[t]

\centering
\caption{Primary matched outcome. Each seed value averages the width-60 cosines ending at rollouts 240, 300, and 360. The interval uses the three paired seed means. Difference means critic--GAE minus group-relative.}
\label{tab:matched}
\begin{tabular}{lrrr}
\toprule
Seed & Group-relative & Critic--GAE & Difference \\
\midrule
42 & $+0.3178$ & $+0.0193$ & $-0.2985$ \\
43 & $+0.2363$ & $+0.0094$ & $-0.2269$ \\
44 & $+0.2229$ & $+0.0060$ & $-0.2168$ \\
\midrule
Mean & $+0.2590$ & $+0.0116$ & $-0.2474$ \\
95\% interval & $[+0.1314,+0.3866]$ & $[-0.0056,+0.0287]$ & $[-0.3580,-0.1368]$ \\
\bottomrule
\end{tabular}
\end{table}

\begin{revisedblock}
The critic mean is close to zero, and its own seed-level interval, $[-0.0056,+0.0287]$, includes zero. Replacing group-relative advantages with the critic--GAE package therefore lowers mean alignment toward zero. Neither local arm reproduces the consistently negative production pattern.

The effect persists in the secondary matched-width summaries. At widths 10, 20, and 40, all paired late-training differences are negative (Appendix~\ref{app:budgets}). Their magnitudes change with width, illustrating why equal aggregation units matter. The group-relative seed mean increases from $0.071$ at width 10 to $0.259$ at width 60. Matching aggregation width makes the measured advantage-system effect interpretable at a shared response and optimizer-update budget.

Training-health checks accompany the geometry measurements. The runs record realized response lengths, EOS completion, and finite updates. Training cap-hit fractions range from $0.213\%$ to $1.189\%$. The fixed 32-example held-out diagnostic shows no consistent improvement over the initial policy. Appendix~\ref{app:protocol} reports the per-run counts. The geometry measurements show sensitivity to the advantage-system configuration, while the held-out diagnostic shows no consistent performance improvement.

\subsection{Numerical accuracy depends on the reported functional}
\label{sec:svdresults}

The SVD validation supports accurate leading-value estimates in the selected sample but exposes material errors in threshold and angle interpretations. With an FP64 Frobenius denominator, the baseline randomized approximation's maximum relative stable-rank error is $8.78\times10^{-5}$, or $0.0088\%$. Including the original FP32 norm and denominator regularizer raises the maximum original-pipeline error to $2.38\%$. The maximum absolute $E_{512}$ error with the FP64 denominator is $0.01353$, or 1.353 percentage points. Appendix~\ref{app:svd} gives the per-matrix errors.

An energy error of this size matters near a reporting threshold. For the \sft{} layer-16 query update, full SVD gives $k_{0.8}=255$ and $E_{256}=0.8002792$. The baseline rank-256 approximation gives $0.7975442$. All three baseline seeds and the stronger randomized setting fall below $0.8$ when normalized by the accurate Frobenius norm. These are four configuration disagreements on one matrix. They demonstrate why an unreached approximate threshold is not a certified lower bound on energy rank.

Subspace angles are more sensitive still. The largest absolute discrepancy between the original FP32 angle calculation and the full-SVD reference is $3.09$ degrees in this sample. For the PPO embedding, the original estimate is $0.09279$ degrees and the full-SVD value is $0.00103$ degrees. Randomized SVD supports efficient broad scans, while these functional-specific errors motivate full decompositions for the reported energy ranks and subspace angles. These findings highlight the importance of validating numerical approximations against the specific spectral or geometric quantity being reported.
\end{revisedblock}

\section{Discussion and Limitations}
\label{sec:limitations}

\begin{revisedblock}
The spectral and trajectory results describe complementary aspects of adaptation. Stable rank measures concentration around the leading singular value, while energy-threshold rank measures the breadth of the spectrum needed to retain a specified energy fraction. Consecutive-displacement cosine adds information about how successive changes interact. The public trajectories show anti-alignment across several RLVR recipes, and the controlled study demonstrates that the advantage system affects alignment magnitude under a shared actor schedule.

The rank reconciliation explains how the reported spectral descriptions coexist. The power-law construction establishes this compatibility analytically, and the full decompositions demonstrate it on selected updates. Its practical implication is that a small stable rank can accompany substantial energy outside a low-rank approximation. Assessing the task-level value of that energy requires evaluating models after spectral truncation.

The public survey measures aggregated changes on each release's native checkpoint grid. Differences in schedules, exposure budgets, storage precision, and related stages limit causal comparisons across releases. The supervised study spans four model families under one data mixture and schedule, with the FP32 follow-up covering two Llama seeds. Its negative Mistral and FP32 Llama pairs show why trajectory sign should be interpreted together with the model and numerical training configuration. The controlled RL study covers one 1B initialization, one task source, one shared trainer, and three paired seeds. Its seed-level intervals characterize variation within that setting.

The controlled intervention combines critic fitting, GAE, and advantage normalization, so its effect applies to that package. Generated-token exposure remains policy-dependent, and the small held-out diagnostic shows no consistent performance improvement. Both local arms have higher mean alignment than the public RLVR trajectories. Identifying the source of this difference requires interventions within the production implementation, with matched exposure and adequate task evaluation.

Full-SVD validation covers fourteen matrices selected by shape and layer position. The resulting accuracy measurements apply to this sample. All geometric quantities describe stored checkpoint weights, whose numerical precision affects apparent support sparsity and small displacements.
\end{revisedblock}

\section{Conclusion}
\label{sec:conclusion}


\begin{revisedblock}
In this work, we studied language-model post-training by separating the structure of cumulative weight updates from the directions of successive training displacements. Our power-law construction showed that low stable rank and large fixed-energy rank are mathematically compatible, and full decompositions verified this separation in selected OLMo-2-7B updates. A dominant singular value can therefore coexist with hundreds of components needed to retain most of an update's energy. This distinction clarifies why different rank measures can give contrasting descriptions of the same matrix.
Trajectory measurements add information about how successive weight changes interact. All 131 measured adjacent pairs across nine public RLVR stages from the OLMo-2, OLMoE, and Olmo-3 releases had negative cosine on their native checkpoint grids, including PPO and GRPO releases. Supervised comparisons of OLMo-2-7B, Llama-3.2-3B, Qwen2.5-3B, and Mistral-7B-v0.3 showed configuration-dependent alignment, with late anti-alignment in Mistral and FP32 Llama runs. In a three-seed matched RL study initialized from OLMo-2-0425-1B-DPO and trained on the RLVR-GSM mathematical-answer dataset, replacing group-relative advantages with the critic--GAE package consistently lowered alignment under a shared actor schedule.
Together, these findings support a more precise account of post-training geometry. Spectral concentration, energy coverage, and trajectory direction answer different questions and should be reported separately. Comparisons also require explicit initialization checkpoints, aggregation widths, and numerical settings. Our full-SVD checks further showed why approximation accuracy must be assessed for the particular quantity being reported. These distinctions provide a basis for interpreting weight-space measurements across post-training procedures.
\end{revisedblock}

\section*{Acknowledgments}
This work was supported by the NVIDIA Academic Grant Program, which provided 8× NVIDIA A100 80 GB GPU resources on the Brev platform.

The authors would like to acknowledge the use of Grammarly (an AI-assisted editing tool), for language editing, proofreading, and typographical corrections during the preparation of this manuscript. All technical content, analyses, and conclusions were manually drafted and verified by the authors.

\section*{Reproducibility and Broader Impact}

\revised{Appendices~\ref{app:protocol}--\ref{app:e4} specify the controlled protocol, exposure units, numerical checks, corrections, and cross-family supervised comparisons.} The accompanying revision artifacts retain per-case results and checkpoint identifiers. No new human-subject data are collected. The study analyzes open-weight models and mathematical-reasoning data. General risks associated with language models remain applicable.

\begingroup\color{black}

\endgroup

\clearpage
\appendix

\section{Proof and Support-Sparsity Examples}
\label{app:proof}

Write $S_k=\sum_{i=1}^ki^{-2\alpha}$. Because $\sigma_1=1$, $\srank=S_n$ and $k_\tau=\min\{k\mid S_k\geq\tau S_n\}$. For $0<\alpha<1/2$, set $b=1-2\alpha$. Integral comparison gives
\[
 S_n\leq1+\int_1^n x^{-2\alpha}\,dx
 =1+\frac{n^b-1}{b},\qquad
 S_n\geq\int_1^{n+1}x^{-2\alpha}\,dx\geq\frac{n^b-1}{b}.
\]
Also $S_k\leq\int_0^k x^{-2\alpha}\,dx=k^b/b$. Hence $S_k\geq\tau S_n$ implies
\[
 k^b\geq\tau(n^b-1),\qquad
 k\geq\tau^{1/b}(n^b-1)^{1/b}.
\]
The latter expression grows linearly with $n$. Since $k_\tau\leq n$, $k_\tau=\Theta(n)$, whereas $\srank=O(n^b)$ and $b<1$. Thus $\srank/k_\tau=O(n^{-2\alpha})\to0$.

For $\alpha=1/2$, $S_k=H_k$ is the harmonic sum, with $\log(k+1)\leq H_k\leq1+\log k$. Consequently $\srank\leq1+\log n$. If $H_k\geq\tau H_n$, then $1+\log k\geq\tau\log n$, so $k\geq n^\tau/e$. This proves $\srank/k_\tau=O(\log n/n^\tau)\to0$ for every fixed $\tau\in(0,1)$ and completes Proposition~\ref{prop:decouple}.

For an illustrative finite dimension, $n=4096$, $\alpha=0.4$, and $\tau=0.8$ give the relaxed upper bound $\srank\leq1+4096^{0.2}/0.2<27.4$ and the lower bound $k_{0.8}\gtrsim469$. This finite-dimensional example illustrates the separation in the constructed spectrum.

The support examples do not require a power law. Let $D=\operatorname{diag}(d_1,\ldots,d_n)$ with every $d_i>0$. Its support always contains exactly $n$ entries, while its singular values are the $d_i$ in decreasing order. Setting all $d_i=1$ gives stable rank $n$; setting $d_1=1$ and the others to a sufficiently small positive value gives stable rank arbitrarily close to one and $k_\tau=1$ for fixed $\tau<1$. For generic orthogonal $U,V$, the matrix $UDV^\top$ has the same spectrum and dense support. Conversely, dense unit vectors $u,v$ give a dense rank-one matrix $uv^\top$. These examples show that support sparsity and spectral concentration describe distinct matrix properties.

\section{Controlled Training Protocol and Health Checks}
\label{app:protocol}

\begin{revisedblock}

\subsection{Shared configuration}

Both arms initialize the policy and frozen reference from \texttt{allenai/OLMo-2-0425-1B-DPO}. Training examples come from \texttt{allenai/RLVR-GSM}, a verifiable mathematical-answer task source related to GSM8K \citep{cobbe2021gsm8k}. We use the same prompt template, numeric final-answer checker, prompt stream, sampling configuration, and actor optimizer configuration within each seed pair. Table~\ref{tab:config} lists the implementation settings.
\end{revisedblock}

\begin{table}[ht]

\centering
\caption{Configuration of the six controlled runs. A rollout batch is the unit that advances the learning-rate schedule and geometry clock.}
\label{tab:config}
\small
\begin{tabular}{p{.43\linewidth}p{.49\linewidth}}
\toprule
Setting & Value \\
\midrule
Paired seeds & 42, 43, 44 \\
Rollout batches & 360 per run \\
Prompts and responses & 8 prompts, 8 responses per prompt \\
Prompt cap and response cap & 2,048 tokens each \\
Sampling & Temperature 1, top-$p$ 1, no top-$k$ restriction \\
Actor update passes & 4 per rollout batch \\
Actor minibatch and microbatch & 8 responses and 1 response \\
Actor optimizer updates & 32 per rollout, 11,520 per run \\
Policy parameters and optimizer states & FP32 with BF16 autocast \\
Optimizer & AdamW, zero weight decay, $\epsilon_{\mathrm{Adam}}=10^{-8}$ \\
Actor learning rate & $4\times10^{-7}$, linear decay over rollouts \\
Ratio clip and gradient norm limit & 0.2 and 1 \\
Reference KL penalty coefficient & 0.05 \\
Reward & Correct final answer 1, incorrect 0, missing EOS $-10$ \\
Critic architecture & Separate model backbone and zero-initialized scalar head \\
Critic optimizer & Separate AdamW, learning rate $4\times10^{-7}$ \\
GAE discount and trace parameter & $\gamma=1$, $\lambda=0.95$ \\
Value clipping & 0.2 around the rollout value estimate \\
Full policy checkpoints & Every 60 rollouts and at completion \\
Online geometry widths & 10, 20, 40, 60 rollout batches \\
\bottomrule
\end{tabular}
\end{table}

\begin{revisedblock}
Each response is trimmed after its first EOS token. That EOS remains in the loss, and subsequent padding does not. Within an eight-response minibatch, we sum token losses and divide by the total retained response-token count. Microbatch accumulation therefore preserves the same token-weighted minibatch objective in both arms. The reference penalty is the nonnegative sample estimator $\exp(z_j)-z_j-1$, where $z_j=\log\pi_{\mathrm{ref}}(y_j\mid x,y_{<j})-\log\pi_\theta(y_j\mid x,y_{<j})$.

The group-relative arm assigns all tokens in response $i$ the normalized outcome advantage $(R_i-\bar R)/(s_R+10^{-6})$, where the mean and population standard deviation are over the eight responses to that prompt. The critic--GAE arm instead forms terminal-reward temporal-difference residuals
\[
 \delta_j=r_j+\gamma V(s_{j+1})-V(s_j),\qquad
 \widehat A_j=\sum_{l=0}^{T-j}(\gamma\lambda)^l\delta_{j+l},
\]
Here $r_j$ denotes the token reward, with the outcome reward assigned to the terminal token and zero reward at earlier tokens; the terminal bootstrap value is zero. The arm whitens advantages over all retained response tokens in the rollout. The critic target is the unwhitened GAE return. Its loss is one half the maximum of unclipped and clipped squared value errors, token averaged within the minibatch. The critic has a separate optimizer, so this experiment changes an advantage-system package that includes critic fitting and normalization.

The shared implementation checks finite losses and gradients and validates saved parameters. The experiment ran on the existing eight-A100 node, with six independent training processes for the paired seeds. The protocol was written before the corrected runs. An execution-order amendment started seeds 43 and 44 concurrently after smoke tests instead of waiting for the seed-42 60-rollout checkpoint. The outcome definitions, hyperparameters, and acceptance criteria remained unchanged, and all completed seeds enter the analysis.

\subsection{Scope relative to published recipes}

Table 18 of the expanded OLMo-2 technical report lists a 2,048-token response cap, learning rate $4\times10^{-7}$ with a linear schedule, and KL coefficient $0.05$ \citep{olmo2025olmo2}. The table is available in the \href{https://arxiv.org/html/2501.00656v3#S5.T18}{expanded technical report}. Our controlled study uses these settings but a 1B model and a smaller rollout batch. The shared custom trainer and critic initialization are specified in Table~\ref{tab:config}. Public release step labels are retained in their original units.

The 2,048-token cap expands the response budget beyond the earlier local 384/768-token experiments. Table~\ref{tab:health} reports the realized cap-hit fractions for the controlled runs.

\subsection{Training health and small held-out diagnostic}

Each run uses the same fixed 32 held-out examples with greedy decoding and the 2,048-token response cap. The initial policy answers 20 of 32 correctly. Table~\ref{tab:health} reports final counts and training cap-hit fractions. The final counts range from 16 to 20 correct answers, compared with 20 initially. This small diagnostic shows no consistent improvement.
\end{revisedblock}

\begin{table}[ht]

\centering
\caption{Health diagnostics for the matched runs. Cap-hit fraction counts training responses that reach 2,048 tokens without EOS. Held-out results use 32 fixed examples and are descriptive.}
\label{tab:health}
\begin{tabular}{llrrr}
\toprule
Arm & Seed & Final correct & Final EOS & Training cap hit \\
\midrule
Group-relative & 42 & 16/32 & 32/32 & 1.102\% \\
Group-relative & 43 & 18/32 & 32/32 & 1.189\% \\
Group-relative & 44 & 20/32 & 32/32 & 0.786\% \\
Critic--GAE & 42 & 19/32 & 27/32 & 0.291\% \\
Critic--GAE & 43 & 18/32 & 28/32 & 0.269\% \\
Critic--GAE & 44 & 18/32 & 29/32 & 0.213\% \\
\bottomrule
\end{tabular}
\end{table}

\section{Exposure Budgets and Matched-Width Results}
\label{app:budgets}

\begin{revisedblock}

All six runs generate 23,040 responses and perform 11,520 actor optimizer updates. Actual generated-token totals differ across policies and seeds, as Table~\ref{tab:budgets} records. These totals count each generated token once across the four update passes. The saved results also retain token counts for every aggregation window.
\end{revisedblock}

\begin{table}[ht]

\centering
\caption{Realized exposure in the matched runs. Every row contains 360 rollouts, 23,040 responses, and 11,520 actor optimizer updates.}
\label{tab:budgets}
\begin{tabular}{llr}
\toprule
Arm & Seed & Generated tokens \\
\midrule
Group-relative & 42 & 5,286,865 \\
Critic--GAE & 42 & 5,832,426 \\
Group-relative & 43 & 4,779,950 \\
Critic--GAE & 43 & 5,861,391 \\
Group-relative & 44 & 5,072,345 \\
Critic--GAE & 44 & 4,776,819 \\
\bottomrule
\end{tabular}
\end{table}

\begin{revisedblock}
For a width $h$, the paired comparison matches $64h$ responses and $32h$ actor optimizer updates. The late-window rule retains cosine endpoints greater than 180. At widths 10, 20, 40, and 60 this gives 18, 9, 5, and 3 cosines per seed, respectively. The width-40 grid includes a window crossing the midpoint, under the prespecified endpoint-based rule. The secondary intervals in Table~\ref{tab:widths} use the same three seed pairs at each width. These descriptive intervals are unadjusted for multiple comparisons across widths.
\end{revisedblock}

\begin{table}[ht]

\centering
\caption{Late-window means and paired uncertainty at matched local widths. Seed values are ordered 42, 43, 44. Intervals are seed-level Student intervals.}
\label{tab:widths}
\small
\begin{tabular}{rlll}
\toprule
Width & Group-relative seeds & Critic--GAE seeds & Paired 95\% interval \\
\midrule
10 & $+0.0897$, $+0.0669$, $+0.0576$ & $+0.0220$, $+0.0140$, $+0.0223$ & $[-0.0922,-0.0118]$ \\
20 & $+0.1392$, $+0.1007$, $+0.0904$ & $+0.0063$, $+0.0093$, $+0.0090$ & $[-0.1698,-0.0341]$ \\
40 & $+0.2594$, $+0.1996$, $+0.1762$ & $+0.0132$, $+0.0121$, $+0.0049$ & $[-0.2997,-0.1037]$ \\
60 & $+0.3178$, $+0.2363$, $+0.2229$ & $+0.0193$, $+0.0094$, $+0.0060$ & $[-0.3580,-0.1368]$ \\
\bottomrule
\end{tabular}
\end{table}

\section{Fixed-Width Temporal Checks and Checkpoint Inventory}
\label{app:temporal}

\begin{revisedblock}

Table~\ref{tab:alllags} reports correlations at separations one, two, and three from full-parameter displacement Gram matrices. Each public or local RL row has six native displacements and therefore five, four, and three pairs. The SFT row has ten displacements and nine, eight, and seven pairs. Each summary averages the available pairs within its trajectory.
\end{revisedblock}

\begin{table}[ht]

\centering
\caption{Fixed-width directional correlations. Local controlled widths are 60 rollouts, the public PPO width is 60 release steps, and SFT width is 500 trainer steps. Only the local controlled arms have matched exposure units.}
\label{tab:alllags}
\small
\begin{tabular}{lrrr}
\toprule
Trajectory & Separation 1 & Separation 2 & Separation 3 \\
\midrule
Public 7B PPO & $-0.220162$ & $-0.024420$ & $-0.008670$ \\
7B SFT seed 42 & $+0.227479$ & $+0.173982$ & $+0.141078$ \\
Group-relative seed 42 & $+0.394134$ & $+0.297558$ & $+0.227189$ \\
Group-relative seed 43 & $+0.349763$ & $+0.251865$ & $+0.210152$ \\
Group-relative seed 44 & $+0.317653$ & $+0.245254$ & $+0.184630$ \\
Critic--GAE seed 42 & $+0.008941$ & $+0.015477$ & $+0.014242$ \\
Critic--GAE seed 43 & $+0.023189$ & $+0.013839$ & $-0.001997$ \\
Critic--GAE seed 44 & $+0.010831$ & $+0.015073$ & $+0.004019$ \\
\bottomrule
\end{tabular}
\end{table}

\begin{revisedblock}
The three OLMo-2-7B SFT seeds are assessed on the same 500-step grid through step 5,000. Their means and sign counts are given below. The longer endpoint trajectory continues the same seed-42 run. Appendix~\ref{app:e4} extends this comparison to the other model families and precision settings.
\end{revisedblock}
\begin{center}

\begin{tabular}{lrr}
\toprule
SFT seed & Mean cosine & Positive pairs \\
\midrule
42 & $+0.227479$ & 9/9 \\
43 & $+0.227273$ & 9/9 \\
44 & $+0.226147$ & 9/9 \\
\bottomrule
\end{tabular}
\end{center}

\begin{revisedblock}
The public 7B PPO calculation includes the immediate DPO initialization and six released checkpoints at labels 60 through 360. Its full-parameter Gram matrix covers 7,298,617,344 stored parameter coordinates. Adjacent cosines are $-0.14379$, $-0.21128$, $-0.23898$, $-0.25304$, and $-0.25372$. Equation~\ref{eq:projection} combines these cosines with displacement-norm ratios to obtain the projection coefficients. Across the nine public stages, the per-stage coefficient ranges are in Table~\ref{tab:projection}.
\end{revisedblock}

\begin{table}[ht]

\centering
\caption{Projection coefficient onto the preceding displacement at native public widths. A negative coefficient describes one component of a step.}
\label{tab:projection}
\begin{tabular}{lrr}
\toprule
Stage & Minimum coefficient & Maximum coefficient \\
\midrule
OLMo-2 7B & $-0.2204$ & $-0.1741$ \\
OLMo-2 13B RLVR1 & $-0.2534$ & $-0.1807$ \\
OLMo-2 13B RLVR2 & $-0.2619$ & $-0.1848$ \\
OLMo-2 13B final & $-0.2643$ & $-0.1837$ \\
OLMo-2 1B RLVR1 & $-0.2779$ & $-0.1042$ \\
OLMo-2 1B final & $-0.2762$ & $-0.1774$ \\
OLMo-2 32B & $-0.2603$ & $-0.1172$ \\
OLMoE 1B active / 7B & $-0.2751$ & $-0.1884$ \\
Olmo-3 7B & $-0.2615$ & $-0.1930$ \\
\bottomrule
\end{tabular}
\end{table}

\begin{revisedblock}
The release repositories and immediate references used in the survey are listed below. All names have the \texttt{allenai/} organization prefix. The accompanying machine-readable results retain full snapshot identifiers where available. Stage names identify the released training segment within each model lineage.
\end{revisedblock}
\begingroup\small\raggedright
\par\smallskip\noindent\begin{minipage}{\linewidth}\textbf{OLMo-2 7B}\\
Release \href{https://huggingface.co/allenai/OLMo-2-1124-7B-Instruct}{OLMo-2-1124-7B-Instruct}.\\
Immediate reference \href{https://huggingface.co/allenai/OLMo-2-1124-7B-DPO}{OLMo-2-1124-7B-DPO}.\end{minipage}\par\smallskip

\par\smallskip\noindent\begin{minipage}{\linewidth}\textbf{OLMo-2 13B RLVR1}\\
Release \href{https://huggingface.co/allenai/OLMo-2-1124-13B-Instruct-RLVR1}{OLMo-2-1124-13B-Instruct-RLVR1}.\\
Immediate reference \href{https://huggingface.co/allenai/OLMo-2-1124-13B-DPO}{OLMo-2-1124-13B-DPO}.\end{minipage}\par\smallskip

\par\smallskip\noindent\begin{minipage}{\linewidth}\textbf{OLMo-2 13B RLVR2}\\
Release \href{https://huggingface.co/allenai/OLMo-2-1124-13B-Instruct-RLVR2}{OLMo-2-1124-13B-Instruct-RLVR2}.\\
Immediate reference \href{https://huggingface.co/allenai/OLMo-2-1124-13B-Instruct-RLVR1}{OLMo-2-1124-13B-Instruct-RLVR1}.\end{minipage}\par\smallskip

\par\smallskip\noindent\begin{minipage}{\linewidth}\textbf{OLMo-2 13B final}\\
Release \href{https://huggingface.co/allenai/OLMo-2-1124-13B-Instruct}{OLMo-2-1124-13B-Instruct}.\\
Immediate reference \href{https://huggingface.co/allenai/OLMo-2-1124-13B-Instruct-RLVR2}{OLMo-2-1124-13B-Instruct-RLVR2}.\end{minipage}\par\smallskip

\par\smallskip\noindent\begin{minipage}{\linewidth}\textbf{OLMo-2 1B RLVR1}\\
Release \href{https://huggingface.co/allenai/OLMo-2-0425-1B-RLVR1}{OLMo-2-0425-1B-RLVR1}.\\
Immediate reference \href{https://huggingface.co/allenai/OLMo-2-0425-1B-DPO}{OLMo-2-0425-1B-DPO}.\end{minipage}\par\smallskip

\par\smallskip\noindent\begin{minipage}{\linewidth}\textbf{OLMo-2 1B final}\\
Release \href{https://huggingface.co/allenai/OLMo-2-0425-1B-Instruct}{OLMo-2-0425-1B-Instruct}.\\
Immediate reference \href{https://huggingface.co/allenai/OLMo-2-0425-1B-RLVR1}{OLMo-2-0425-1B-RLVR1}.\end{minipage}\par\smallskip

\par\smallskip\noindent\begin{minipage}{\linewidth}\textbf{OLMo-2 32B}\\
Release \href{https://huggingface.co/allenai/OLMo-2-0325-32B-Instruct}{OLMo-2-0325-32B-Instruct}.\\
Immediate reference \href{https://huggingface.co/allenai/OLMo-2-0325-32B-DPO}{OLMo-2-0325-32B-DPO}.\end{minipage}\par\smallskip

\par\smallskip\noindent\begin{minipage}{\linewidth}\textbf{OLMoE 1B active / 7B}\\
Release \href{https://huggingface.co/allenai/OLMoE-1B-7B-0125-Instruct}{OLMoE-1B-7B-0125-Instruct}.\\
Immediate reference \href{https://huggingface.co/allenai/OLMoE-1B-7B-0125-DPO}{OLMoE-1B-7B-0125-DPO}.\end{minipage}\par\smallskip

\par\smallskip\noindent\begin{minipage}{\linewidth}\textbf{Olmo-3 7B}\\
Release \href{https://huggingface.co/allenai/Olmo-3-7B-Instruct}{Olmo-3-7B-Instruct}.\\
Immediate reference \href{https://huggingface.co/allenai/Olmo-3-7B-Instruct-DPO}{Olmo-3-7B-Instruct-DPO}.\end{minipage}\par\smallskip
\endgroup

\begin{table}[ht]

\centering
\caption{Additional matrix-shape inventory for the public releases. Counts cover matrices of the stated shape and exclude one-dimensional normalization weights. The latter remain in full-parameter cosine calculations. OLMoE counts include individual expert matrices and router matrices.}
\label{tab:otherdimensions}
\small
\begin{tabular}{lrrr}
\toprule
Model & Rows & Columns & Count \\
\midrule
OLMo-2 1B final & 100,352 & 2,048 & 2 \\
 & 2,048 & 8,192 & 16 \\
 & 8,192 & 2,048 & 32 \\
 & 2,048 & 2,048 & 64 \\
OLMo-2 13B final & 100,352 & 5,120 & 2 \\
 & 5,120 & 13,824 & 40 \\
 & 13,824 & 5,120 & 80 \\
 & 5,120 & 5,120 & 160 \\
OLMo-2 32B & 100,352 & 5,120 & 2 \\
 & 5,120 & 27,648 & 64 \\
 & 27,648 & 5,120 & 128 \\
 & 1,024 & 5,120 & 128 \\
 & 5,120 & 5,120 & 128 \\
Olmo-3 7B & 100,278 & 4,096 & 2 \\
 & 4,096 & 11,008 & 32 \\
 & 11,008 & 4,096 & 64 \\
 & 4,096 & 4,096 & 128 \\
OLMoE 1B active / 7B & 50,304 & 2,048 & 2 \\
 & 2,048 & 1,024 & 1,024 \\
 & 1,024 & 2,048 & 2,048 \\
 & 64 & 2,048 & 16 \\
 & 2,048 & 2,048 & 64 \\
\bottomrule
\end{tabular}
\end{table}

\section{Full-SVD Validation}
\label{app:svd}

\begin{revisedblock}

The validation uses NumPy 1.26.4, scikit-learn 1.8.0, and PyTorch 2.12.0+cu130. A reduced QR factorization followed by LAPACK full SVD preserves the complete singular spectrum for rectangular cases. The largest selected matrix is the $100352\times4096$ embedding, for which reduced QR makes the full-spectrum calculation tractable.

For subspaces, let $U_0,U_T$ contain the leading 64 left singular vectors of the reference and endpoint weight matrices. We compute the largest principal angle as $\arcsin\|(I-U_0U_0^\top)U_T\|_2$ using orthonormal FP64 bases. This avoids the small-angle floor of an arccos calculation. We separately reproduce the original FP32 arccos calculation. Table~\ref{tab:svderrors} gives baseline approximation errors and both angle estimates.

\emph{Original angle} is the largest principal angle computed by the original randomized-SVD and FP32 procedure. Let $\widehat U_0,\widehat U_T$ be the first 64 left singular vectors returned by the rank-256 approximations. In degrees, this calculation is
\[
 \phi_{\mathrm{original}}=\frac{180}{\pi}\arccos\!\left(
 \operatorname{clip}_{[-1,1]}\!\left[\sigma_{\min}(\widehat U_0^\top\widehat U_T)\right]\right).
\]
The cross-product and its singular values are computed in FP32. For energy, the approximation and absolute error are
\[
 \widehat E_{512}=\frac{\sum_{i=1}^{512}\widehat\sigma_i^2}{\|\Delta\|_F^2},
 \qquad
 \varepsilon_E=\left|\widehat E_{512}-E_{512}^{\mathrm{full}}\right|.
\]
Here $\widehat\sigma_i$ are the rank-512 randomized-SVD singular values, $E_{512}^{\mathrm{full}}$ comes from the full decomposition, and both fractions use the same FP64 squared Frobenius norm. Table~\ref{tab:svderrors} reports $\varepsilon_E$ as a fraction; $0.01$ equals one percentage point.
\end{revisedblock}

\begin{table}[ht]

\centering
\caption{Baseline numerical audit on the fourteen selected matrices. The stable-rank error isolates singular-value error with an FP64 Frobenius norm. Energy error is absolute error in $E_{512}$. Here $E_{512}=E_k(\Delta)$ from Equation~\ref{eq:ranks} at $k=512$; the energy-error and original-angle formulas appear above. Angles are the largest of the 64 principal angles, in degrees. The original-pipeline stable-rank error can be larger because its normalization differs.}
\label{tab:svderrors}
\scriptsize
\begin{tabular}{llrrrr}
\toprule
Run & Matrix & Relative stable-rank error & Energy error & Original angle & Full-SVD angle \\
\midrule
PPO & Query 0 & 3.9e-07 & 0.00514 & 0.31404 & 0.29726 \\
PPO & MLP up 0 & 1.5e-06 & 0.00939 & 0.09487 & 0.07653 \\
PPO & Query 16 & 5.7e-07 & 0.00818 & 0.09691 & 0.05868 \\
PPO & MLP down 16 & 4.3e-07 & 0.01029 & 0.08623 & 0.08062 \\
PPO & Query 31 & 1e-07 & 0.00761 & 0.13848 & 0.13671 \\
PPO & MLP up 31 & 1.7e-07 & 0.01044 & 0.13122 & 0.17695 \\
PPO & Embedding & 8.8e-05 & 0.01350 & 0.09279 & 0.00103 \\
SFT & Query 0 & 4.2e-07 & 0.00029 & 76.20685 & 76.26711 \\
SFT & MLP up 0 & 1.1e-06 & 0.01247 & 1.62522 & 1.96539 \\
SFT & Query 16 & 2e-07 & 0.00296 & 8.71772 & 8.72792 \\
SFT & MLP down 16 & 7.5e-05 & 0.01282 & 1.94431 & 2.76619 \\
SFT & Query 31 & 2e-07 & 0.00412 & 5.86628 & 5.85589 \\
SFT & MLP up 31 & 3.9e-07 & 0.01231 & 5.37735 & 2.28540 \\
SFT & Embedding & 1.9e-05 & 0.01353 & 0.10279 & 0.06163 \\
\bottomrule
\end{tabular}
\end{table}

\begin{revisedblock}
The threshold failure occurs in the SFT layer-16 query projection. Full SVD gives $E_{256}=0.80027920$ and $k_{0.8}=255$. Independent baseline rank-256 calls at seeds 42, 43, and 44 give $0.79754418$, $0.79760081$, and $0.79759122$. Forty oversamples and eight iterations at seed 42 give $0.79972833$. All four approximate energies use the same FP64 denominator as the full result. The original FP32 denominator introduces an additional shift. Full-spectrum ranks resolve the threshold for these selected cases.
\end{revisedblock}

\section{Corrections and Disposition of Earlier Analyses}
\label{app:corrections}

\begin{revisedblock}
The original public 7B PPO endpoint subtracted SFT and therefore contained both DPO and PPO changes. Revised PPO spectral cases subtract the immediate DPO checkpoint. We withdraw the earlier MLP stable-rank headline near 37 as a pure-PPO statistic. Separately, the intended reservoir sketch admitted no coordinates after its first two million entries. All reported full-model cosines now use full-coordinate accumulation, and we withdraw the earlier sketch-error estimate.

The reference cancels between two fixed checkpoints but affects the initial displacement and all cumulative descriptors. Independently, correcting the sketch can change every previously sampled global cosine.

The Qwen ablations had unequal optimizer-update counts, unset critic seeds, and post-EOS padding entering losses. We replace them with the shared, paired implementation in Section~\ref{sec:matchedsetup} and withdraw claims that clipping is ruled out or that an advantage estimator explains production anti-alignment.

The original local 7B GRPO run started from SFT, not DPO, and its requested 60-step reconstruction is unavailable in the retained inventory. We remove the matched-initialization claim. The paired 1B experiment supplies the controlled comparison in the revised analysis. Energy-rank and angle claims follow Section~\ref{sec:svdresults}. These corrections preserve the spectral compatibility proposition.
\end{revisedblock}

\begin{revisedblock}
\section{Cross-Family SFT and Numerical Precision}
\label{app:e4}

\subsection{Shared training and measurement protocol}
\label{sec:e4protocol}

The cross-family study extends the OLMo-2-7B comparison to \path|meta-llama/Llama-3.2-3B|, \path|Qwen/Qwen2.5-3B|, and \path|mistralai/Mistral-7B-v0.3|. All runs train on \path|allenai/tulu-3-sft-olmo-2-mixture| with the same plain-text system, user, and assistant role formatting. The causal language-model loss supervises all retained non-padding conversation tokens with a preceding token, as specified in Equation~\ref{eq:sft}. The maximum sequence length is 4,096 tokens. The effective batch contains 64 examples, and AdamW uses learning rate $10^{-5}$, zero weight decay, 3\% warmup, and linear decay over the full two-epoch schedule of 29,356 optimizer steps. A stop callback ends the comparison at step 5,000 without rescaling that schedule.

The baseline runs hold model parameters and Adam states in BF16. The Llama precision follow-ups hold both in FP32 and perform forward and backward computation under BF16 autocast. The BF16 Llama and Qwen runs use microbatches of two examples per device, while the Mistral and FP32 Llama runs use one; gradient accumulation and device count maintain the effective batch of 64. All arms use gradient checkpointing. The precision comparison consequently concerns the numerical training configuration, including parameter and optimizer-state precision.

At steps 0, 500, \ldots, 5,000, we record the full parameter vector. The checkpoint-based and in-memory implementations both form displacements in FP64 before accumulating their full-coordinate Gram matrix. For tied parameters, each independent parameter is counted once. The in-memory method avoids retaining another complete checkpoint series on disk. A comparison on Mistral seed 43 at steps 0, 500, and 1,000 gives a maximum absolute Gram-entry discrepancy of $1.28\times10^{-13}$ and relative discrepancy of $1.75\times10^{-15}$ across 7,248,023,552 parameter coordinates. Table~\ref{tab:e4} reports every completed 5,000-step trajectory in the study.

The extensions were sequential. Three Llama seeds formed the initial cross-family test; Mistral and Qwen were added subsequently. The negative Mistral outcome motivated a second Mistral seed and the Llama precision follow-ups. A 2,000-step FP32 seed-42 pilot preceded a fresh 5,000-step run with the same seed and schedule. Its first three cosines, $+0.1766$, $+0.0678$, and $+0.0233$, agree closely with the corresponding values in Table~\ref{tab:e4precision}. The tables summarize the complete 5,000-step grids, with the pilot treated as a diagnostic of the same setting.
\end{revisedblock}

\begin{table}[ht]

\centering
\caption{SFT comparisons at width 500 optimizer steps through step 5,000. Precision denotes trainable parameters and Adam states. Each run contains nine adjacent displacement pairs. OLMo rows reproduce the seed summary in Appendix~\ref{app:temporal}.}
\label{tab:e4}
\small
\begin{tabular}{lllrrr}
\toprule
Model & Precision & Seed & Mean cosine & Positive & Negative \\
\midrule
OLMo-2-7B & BF16 & 42 & $+0.2275$ & 9/9 & 0/9 \\
OLMo-2-7B & BF16 & 43 & $+0.2273$ & 9/9 & 0/9 \\
OLMo-2-7B & BF16 & 44 & $+0.2261$ & 9/9 & 0/9 \\
Llama-3.2-3B & BF16 & 42 & $+0.2375$ & 9/9 & 0/9 \\
Llama-3.2-3B & BF16 & 43 & $+0.2381$ & 9/9 & 0/9 \\
Llama-3.2-3B & BF16 & 44 & $+0.2372$ & 9/9 & 0/9 \\
Qwen2.5-3B & BF16 & 42 & $+0.2226$ & 9/9 & 0/9 \\
Mistral-7B-v0.3 & BF16 & 42 & $-0.0100$ & 2/9 & 7/9 \\
Mistral-7B-v0.3 & BF16 & 43 & $-0.0099$ & 2/9 & 7/9 \\
Llama-3.2-3B & FP32 & 42 & $+0.0206$ & 4/9 & 5/9 \\
Llama-3.2-3B & FP32 & 43 & $+0.0209$ & 4/9 & 5/9 \\
\bottomrule
\end{tabular}
\end{table}

\begin{revisedblock}
\subsection{Family and precision dependence}
\label{sec:e4results}

The BF16 Llama comparison reproduces positive alignment across three seeds, with means $+0.2375$, $+0.2381$, and $+0.2372$. The Qwen run also has nine positive pairs and mean $+0.2226$. In contrast, the two Mistral seeds each have seven negative pairs, beginning with the pair ending at step 2,000. Their full-grid means are $-0.0100$ and $-0.0099$. Thus the shared supervised recipe produces different directional patterns across models.

The temporal summaries retain this contrast at larger widths. Table~\ref{tab:e4lags} holds width fixed when increasing separation and separately aggregates displacements into widths 1,000 and 1,500. The BF16 Llama and Qwen means remain positive under these summaries. Mistral's width-1,000 means are approximately $-0.0159$ for both seeds.

Within Llama, changing the numerical training configuration substantially lowers alignment. The FP32 seed-42 and seed-43 full-grid means are $+0.0206$ and $+0.0209$, with four positive and five negative pairs each. Table~\ref{tab:e4precision} shows the change over training. Both runs first become negative at the pair ending at step 3,000 and remain negative through step 5,000. Full-grid means remain positive because of the early positive pairs. These results establish that the positive sign in the BF16 Llama runs is sensitive to numerical training configuration; they also show that supervised updates can exhibit consecutive anti-alignment.
\end{revisedblock}

\begin{table}[ht]

\centering
\caption{Additional cross-family SFT temporal summaries. Separation-two and separation-three means use width 500 and contain eight and seven pairs. Width-1,000 and width-1,500 means aggregate consecutive displacements and contain four and two adjacent pairs. The width-1,500 grid ends at step 4,500.}

\label{tab:e4lags}
\small
\begin{tabular}{lllrrrr}
\toprule
Model & Precision & Seed & Sep.\ 2 & Sep.\ 3 & Width 1,000 & Width 1,500 \\
\midrule
Llama-3.2-3B & BF16 & 42 & $+0.1889$ & $+0.1593$ & $+0.3195$ & $+0.3609$ \\
Llama-3.2-3B & BF16 & 43 & $+0.1891$ & $+0.1578$ & $+0.3194$ & $+0.3594$ \\
Llama-3.2-3B & BF16 & 44 & $+0.1881$ & $+0.1584$ & $+0.3172$ & $+0.3605$ \\
Qwen2.5-3B & BF16 & 42 & $+0.1784$ & $+0.1510$ & $+0.3037$ & $+0.3449$ \\
Mistral-7B-v0.3 & BF16 & 42 & $-0.0031$ & $-0.0005$ & $-0.0159$ & $-0.0144$ \\
Mistral-7B-v0.3 & BF16 & 43 & $-0.0035$ & $-0.0004$ & $-0.0159$ & $-0.0150$ \\
Llama-3.2-3B & FP32 & 42 & $+0.0273$ & $+0.0275$ & $+0.0408$ & $+0.0691$ \\
Llama-3.2-3B & FP32 & 43 & $+0.0269$ & $+0.0273$ & $+0.0408$ & $+0.0683$ \\
\bottomrule
\end{tabular}
\end{table}

\begin{table}[ht]

\centering
\caption{Llama precision comparison at width 500. Each entry is the cosine between the displacement ending at the listed step and its immediate predecessor. BF16 and FP32 denote parameter and Adam-state precision.}
\label{tab:e4precision}
\small
\begin{tabular}{rrrrr}
\toprule
& \multicolumn{2}{c}{Seed 42} & \multicolumn{2}{c}{Seed 43} \\
\cmidrule(lr){2-3}\cmidrule(lr){4-5}
Endpoint & BF16 & FP32 & BF16 & FP32 \\
\midrule
1000 & $+0.3724$ & $+0.1765$ & $+0.3676$ & $+0.1774$ \\
1500 & $+0.3763$ & $+0.0678$ & $+0.3767$ & $+0.0708$ \\
2000 & $+0.2993$ & $+0.0232$ & $+0.3023$ & $+0.0226$ \\
2500 & $+0.2454$ & $+0.0031$ & $+0.2478$ & $+0.0031$ \\
3000 & $+0.2120$ & $-0.0066$ & $+0.2072$ & $-0.0095$ \\
3500 & $+0.1868$ & $-0.0142$ & $+0.1911$ & $-0.0141$ \\
4000 & $+0.1648$ & $-0.0184$ & $+0.1660$ & $-0.0177$ \\
4500 & $+0.1493$ & $-0.0217$ & $+0.1481$ & $-0.0216$ \\
5000 & $+0.1309$ & $-0.0242$ & $+0.1359$ & $-0.0231$ \\
\bottomrule
\end{tabular}
\end{table}

\end{document}